\pdfoutput=1

\documentclass{INTERSPEECH2023}

\usepackage{xcolor}
\usepackage{multirow}
\usepackage{comment}
\usepackage{pgfplots,subcaption}
\pgfplotsset{compat=1.17}

\newcommand{\mg}{\textcolor{black}}

\newcommand{\mn}{\textcolor{black}}

\interspeechcameraready

\title{Joint Speech Translation and Named Entity Recognition}
\name{Marco Gaido$^{\dagger}$, Sara Papi$^{\dagger \ddagger}$, Matteo Negri$^{\dagger}$, Marco Turchi$^{\star*}$}
\address{$^{\dagger}$Fondazione Bruno Kessler, Italy
      $^{\ddagger}$University of Trento, Italy 
      $^{\star}$Independent Researcher}
\email{\{mgaido, spapi, negri\}@fbk.eu,marco.turchi@gmail.com}

\begin{document}

\maketitle
 
\begin{abstract}
Modern automatic translation systems aim at supporting the users by providing contextual knowledge. In this framework, a critical task is the output enrichment with information regarding the mentioned entities. This is currently achieved by processing the generated translations with named entity recognition (NER) tools and retrieving their description from knowledge bases. In light of the recent promising results shown by direct speech translation (ST) models and the known weaknesses of cascades (error propagation and additional latency), in this paper we propose multitask models that jointly perform ST and NER, and compare them with a cascade baseline. Experimental results on three language pairs (en-es/fr/it) show that our models significantly outperform the cascade on the NER task (by 0.4-1.0 F1), without degradation in terms of translation quality, and with the same computational efficiency of a plain direct ST model. 
\end{abstract}

\section{Introduction}

Drawing inspiration from augmented reality, where real-world vision is complemented with 
overlaid
relevant information, ``augmented translation'' \cite{lommel-2018-augmented} is an emerging research line aimed to enrich automatically-generated translations with semantic information by highlighting named entities (NEs) and key concepts (the focus of this work) and eventually linking them to external knowledge bases~(an aspect we do not cover here).
On one side, this can ease, speed up, and improve the generation of fluent and high-quality translations by professional translators and post-editors; on the other, it provides end users with additional information that may be needed to fully understand a sentence, especially in highly specialized domains.\footnote{https://intelligent-information.blog/en/augmented-translation-puts-translators-back-in-the-center/}

Current solutions 
rely on a cascade architecture comprising a text-to-text machine translation (MT) system whose output is fed to a NE recognition (NER) model \cite{depalma}. No work has instead explored its application to speech-to-text translation (ST), and the possibility of jointly performing the ST and NER tasks
with a single model, despite positive signals from related fields. Indeed, in the task of NER from speech, the traditional cascade approach -- composed of an automatic speech recognition (ASR) system followed by an NER model -- has been recently challenged by the competitiveness of direct models that perform the two tasks jointly \cite{Ghannay-2018-entoend-ner,yadav20b_interspeech,caubriere-etal-2020-named,boli-2022-aishell-ner}.
Similarly, in 
MT,
multitask models that jointly perform MT and NER have been shown to improve NE accuracy without degrading translation quality \cite{xie-2022-e2e-ea-nmt}.

In light of this and 
the competitive results of direct ST models~\cite{berard_2016,weiss2017sequence} compared to the conventional ASR+MT pipeline~\cite{bentivogli-etal-2021-cascade,anastasopoulos-etal-2021-findings}, in this paper we address 
two
research questions: 
\textbf{(1)} \textit{Is
the current cascade of an ST system (either direct ST or ASR+MT) followed by an NER tool better than performing the two tasks with a single model?}
\textbf{(2)} \textit{What are the effects on NE accuracy and translation quality of
using a single multi-task model?}

To answer these questions, we explore different methods to jointly perform ST and NER.\footnote{Code available at: \url{https://github.com/hlt-mt/FBK-fairseq}.}
Our experiments on three language pairs 
show
that joint models significantly outperform the ST+NER cascade by 0.4-1.0 F1 in the NER task while being on par in terms of translation quality. 
Such improvement is achieved without introducing any 
computational overhead with respect to a plain ST model, making our solution remarkably more efficient than the cascade approach. This is directly reflected in the computational-aware latency in simultaneous ST scenarios, where our best 
model jointly performs ST and NER with the same latency (and quality)
as an ST-only model.


\section{Joint NER and ST}
\label{sec:ner_st}

The easiest way to extract the NEs from a translation consists in applying an NER model on the output of the ST model.
Henceforth, we refer to this approach as \textit{cascade}, and we consider it as a baseline for comparison against our systems that jointly perform the two tasks with a single model. Our solutions -- \textit{inline}, and \textit{parallel} -- are described below:

\begin{figure}[!t]
\centering
\includegraphics[width=0.43\textwidth]{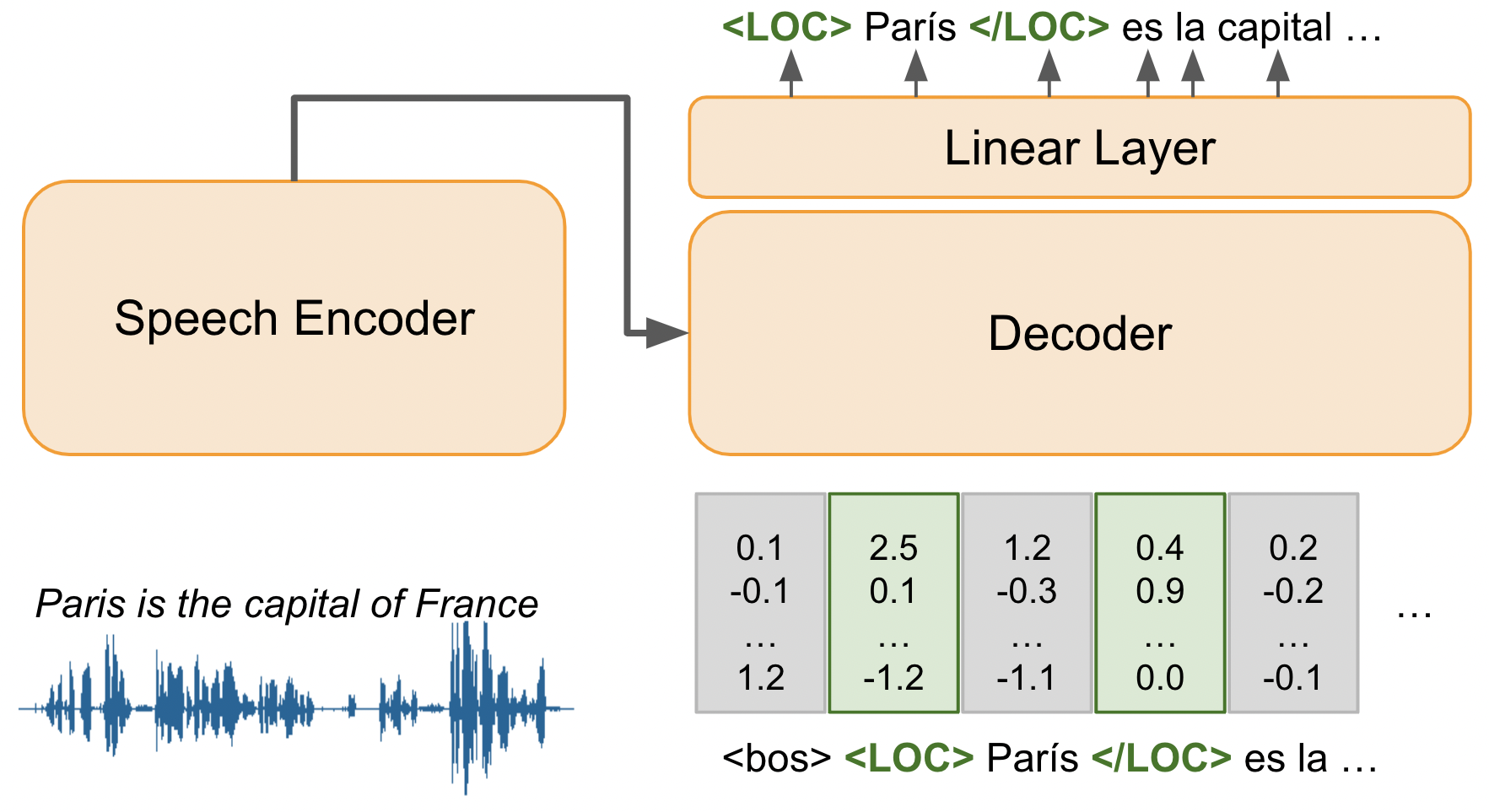}
\caption{\label{fig:inline} Architecture of the \textit{inline} solution. The additional tokens generated in the output are highlighted in green, and are passed to the decoder as  all the other previous output tokens.} 

\end{figure}

\noindent\textbf{Inline (Fig. \ref{fig:inline})}. The vocabulary of the direct ST model is extended with tags that represent the start (e.g., \texttt{<LOC>}) and end (e.g., \texttt{</LOC>}) of the NE categories to be recognized which, in our case, are 18,\footnote{The categories are those defined in the OntoNotes annotation \cite{ontonotes}.} for a total of 36 entries.
These tags are treated as all other tokens (subwords):
they are predicted in the output sequence, and -- together with the other tokens -- fed to the decoder as previous output tokens, informing it about the NE categories.
This solution does not require
architectural changes to the ST model but introduces 
additional overhead, especially at inference time, as the higher number of tokens to generate (due to the additional start/end NE tags) leads to an increase in the number of forward passes on the autoregressive decoder.

\begin{figure}[!t]
\centering
\includegraphics[width=0.415\textwidth]{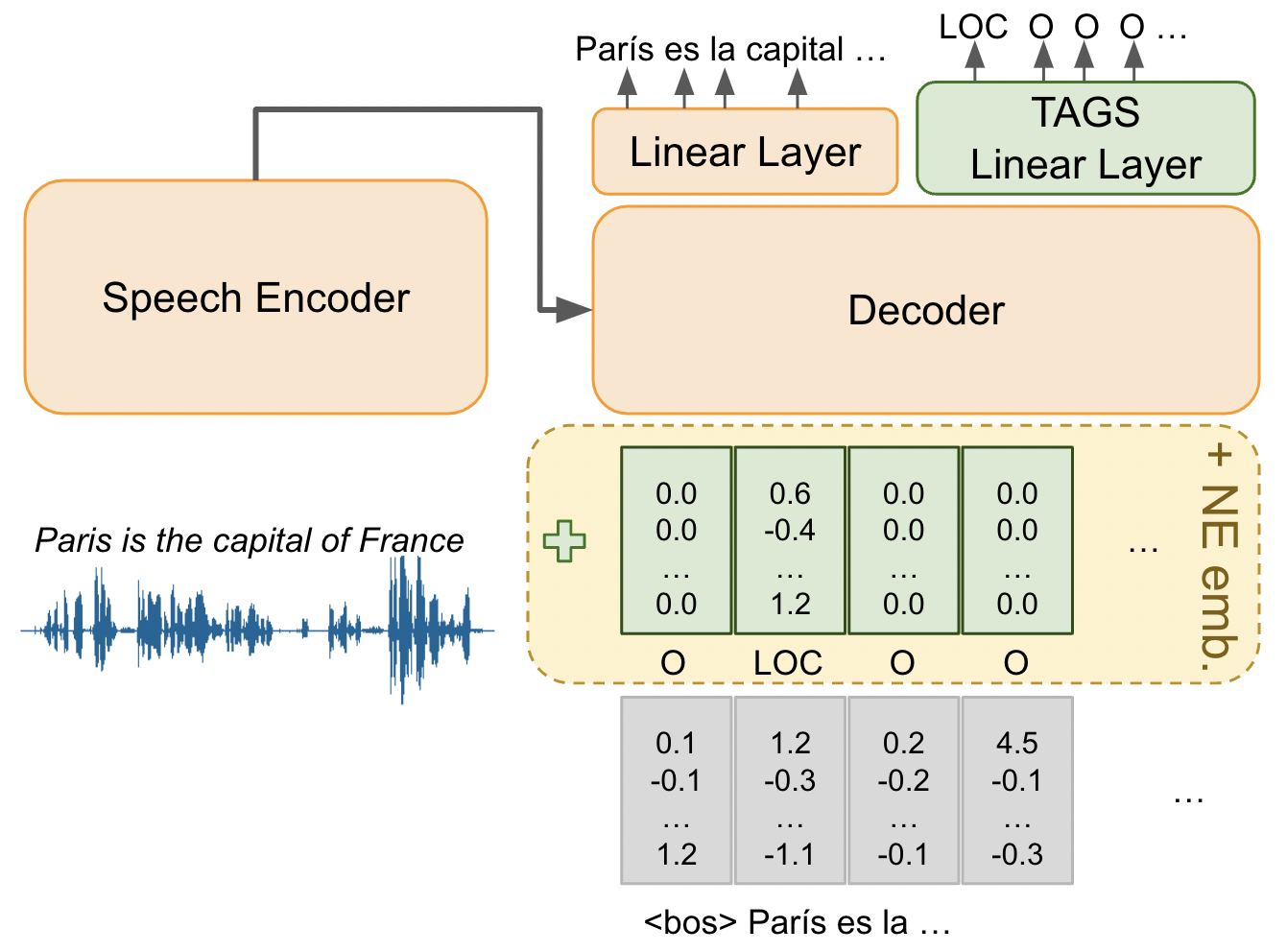}
\caption{\label{fig:parallel} Architecture of the \textit{parallel} solution.
The introduced linear layer (in green) is processed token-by-token in parallel with the other linear layer. 
In the \textit{+ NE emb.} variant (yellow dotted area), the 
\mg{previous tags} are converted into embeddings that are summed to those of the corresponding 
\mg{previous tokens.}}
\end{figure}

\noindent\textbf{Parallel (Fig. \ref{fig:parallel})}. At each time step, two linear layers process in parallel the output of the last decoder layer: one
maps the vectors to the vocabulary space to predict the next token as in standard ST models; the other
maps the same vectors to the NE-category space to predict the NE category to which the token belongs, if any, or \textit{O} (i.e. \textit{OTHER}), if the token is not part of a NE.
Although the second linear layer introduces additional parameters to train, its computational cost is negligible compared to that of the whole decoder. Moreover, this solution avoids the supplementary decoder forward passes required by the inline method.
However, the potential drawback in comparison with the inline solution is that it cannot exploit information about the NE categories
predicted for the previously generated tokens during translation.
As we posit that this lack of information may cause performance degradation, we propose a variant of this method in which the embeddings of the previous output tokens are summed with
learned embeddings of their corresponding NE categories.\footnote{The beginning-of-sentence (\textit{bos}) token is considered of \textit{O} category.}
This change requires only 19 additional embeddings to learn (one for each NE category, plus \textit{O}) --~a negligible number compared to the target vocabulary size~-- and a sum, hence producing no significant computational overhead.
We refer to this variant as \textbf{Parallel + NE emb.}

\section{Experimental Settings}
\label{sec:exp_settings}

\noindent\textbf{\mg{Models.}} All our ST models are fed with 80 features extracted from the audio every 10ms with sample windows of 25ms.
These sequences of features are processed by two 1D convolutional layers that reduce the sequence length by a factor of 4, before passing them to a 12-layers Conformer encoder \cite{gulati20_interspeech}, and a 6-layer autoregressive Transformer decoder \cite{transformer}.
We use 512 features with 1024 hidden neurons in the FFN for both the encoder and decoder. The target vocabulary is created with 8,000 BPE~\cite{sennrich-etal-2016-neural} merge rules.
As a result, our models have 116M parameters that we optimize with label-smoothed cross-entropy loss \cite{szegedy2016rethinking} (0.1 smoothing factor)
and
an auxiliary CTC~\cite{Graves2006ConnectionistTC} loss on the output of 8\textsuperscript{th} encoder layer with the transcript as the target to improve model convergence \cite{9003774}. Moreover, we adopt 
\mn{CTC}
\mg{compression \cite{gaido-etal-2021-ctc} to reduce}
the input dimension and speed up both training and inference.
As optimizer, we use Adam~\cite{adam}, and the learning rate is initially increased for 20k steps up to 
0.005 and then it decreases with the inverse squared root policy.
We train on 4 K80 GPUs with 10k tokens per mini-batch and 8 as update frequency.
The training stops after 5 epochs without loss decrease on the validation set, and average 5 checkpoints around the best. 
At inference time, we decode using beam search with 5 as beam size.
The ASR model of our ASR+MT+NER pipelines
is trained on the same data and with the same method described for the ST models.
We
rely on a multilingual BERT-based model,\footnote{http://docs.deeppavlov.ai/en/master/features/models/bert.html} openly-available in DeepPavlov \cite{burtsev-etal-2018-deeppavlov}, as NER system and, on the 1.3B-parameters distilled NLLB \cite{costa2022no} as MT model.

\noindent\textbf{\mg{Data and Evaluation Metrics.}}
All models are trained on
MuST-C \cite{MuST-Cjournal} and Europarl-ST~\cite{iranzo-2020-europarl-st}.
To train the joint NER and ST models, we automatically annotated the NEs on the target translations with the same NER tool used in our cascade approach, obtaining parallel training data with speech and the corresponding annotated translations without any manual intervention. Translation quality is evaluated with SacreBLEU\footnote{case:mixed$\vert$eff:no$\vert$tok:13a$\vert$smooth:exp$\vert$version:2.0.0} \cite{post-2018-call} on the Europarl-ST test set.
Regarding NEs, instead, we measure three aspects on NEuRoparl-ST benchmark \cite{gaido-etal-2021-moby}: \textit{1)} the generation of the correct translation, \textit{2)} the recognition (or identification) of the NEs in the generated text, and \textit{3)} the classification with the correct NE category. First, we use NE accuracy (case-insensitive, for the sake of comparison with previous work) to assess
the ability to translate NEs. Second, we compute F1 to measure the ability in recognizing NEs, although F1 is also influenced by the NE translation quality, as it is computed by considering as correct only those NEs that are accurately translated and identified, but disregarding their category classification.
As such, NEs that are poorly translated and recognized by a model penalize both recall and precision. 
The strict F1 definition mirrors the users' perception:
in augmented ST, while unrecognized NEs are only a lack of help to the users, recognized but
incorrect
NEs are more harmful as they would distract them with unrelated and potentially misleading content.
Lastly, we use classification accuracy to measure the percentage of NEs assigned to the correct category.


\begin{table*}[!bt]
\setcounter{table}{1}
\centering
\setlength{\tabcolsep}{1.75pt}
\small
\begin{tabular}{l|cccc|cccc|cccc}
\hline
\multirow{2}{*}{\textbf{Model}}
 & \multicolumn{4}{c|}{\textbf{en-es}} & \multicolumn{4}{c|}{\textbf{en-fr}} & \multicolumn{4}{c}{\textbf{en-it}} \\ 
 \cline{2-13}
 & \textbf{BLEU} & \textbf{NE Acc} & \textbf{F1} & \textbf{Cat. Acc.} & \textbf{BLEU} & \textbf{NE Acc} & \textbf{F1} & \textbf{Cat. Acc.} & \textbf{BLEU}  & \textbf{NE Acc} & \textbf{F1} & \textbf{Cat. Acc.} \\
\hline
Prev. work \cite{gaido-etal-2021-moby} & 37.7 & 71.4 & - & - & 30.1 & 67.3 & - & - & 26.0 & 67.3 & - & - \\
\hline
Cascade (ASR+MT+NER) & 37.3 & 71.0 & 47.4 & 89.9 & \textbf{37.6} & 68.8 & 44.6 & 90.2 & 26.2 & 64.8 & 42.0 & 87.5 \\
Cascade (ST+NER) & 37.9 & 71.9 & 49.1 & 89.8 & 36.2 & 69.2 & 44.8 & 90.2 & 28.3 & 66.5 & 44.5 & 88.8 \\
\hline
Inline & 37.9 & \textbf{72.2} & \textbf{49.5}$^{\dagger\ddagger}$ & \textbf{90.1} & 36.3 & \textbf{69.6} & \textbf{45.6}$^{\dagger\ddagger}$ & 90.2 & 28.3 & 66.9 & \textbf{45.5}$^{\dagger\ddagger}$ & \textbf{89.4} \\ 
Parallel & \textbf{38.1} & 71.9 & 48.1 & 89.5 & 36.1 & 69.0 & 44.5 & \textbf{90.6} & \textbf{28.4} & \textbf{67.5} & 43.9 & 89.1 \\
\hspace{2mm} + NE emb. & 38.0 & 72.1 & \textbf{49.5}$^{\dagger\ddagger}$ & 89.9 & 36.1 & 69.3 & 45.5$^{\dagger\ddagger}$ & 90.4 & 28.2 & 67.3 & 45.4$^{\dagger\ddagger}$ & 89.1 \\
\hline
\end{tabular}
\caption{
BLEU ($\uparrow$),
case-insensitive NE accuracy ($\uparrow$), F1 ($\uparrow$), and category
classification accuracy (Cat. Acc., $\uparrow$) of previous work, 
our cascades (ASR+MT+NER and ST+NER)
and the proposed joint
ST\&NER
models. All results are the average of three runs. $^\dagger$ indicates statistically significant improvements over
ST+NER, and $^\ddagger$ over \textit{parallel}. A result is considered statistically significant 
if
we can reject with 95\% confidence the null hypothesis that the considered mean is not higher than the mean of the baseline with Student's t-test \cite{student}.}
\label{tab:results}
\end{table*}

\section{Results}
\label{sec:results}


\noindent\textbf{Translation Quality (Overall and at NE Level).}
First, to ensure the 
soundness of our experimental settings and,
in turn, of our analysis, we compare our base direct ST model with recent works on Europarl-ST (Table \ref{tab:comparison_prev_work}). 
As shown by the results, our systems outperform, to the best of our knowledge, all recently published scores on the same benchmark.
This confirms 
the strength of our models and the reliability of our results.

\begin{table}[!bt]
\setcounter{table}{0}
\centering
\setlength{\tabcolsep}{10pt}
\small
\begin{tabular}{l|ccc}
\hline
\textbf{Model} & \textbf{en-es} & \textbf{en-fr} & \textbf{en-it} \\
\hline
ASR+MT \cite{iranzo-2020-europarl-st} & 28.0 & 23.4 & - \\
NPDA-kNN-ST \cite{du-etal-2022-non} & 29.0 & 27.7 & 20.5 \\
STR+KD \cite{lam-etal-2022-sample} & - & 29.3 & - \\
NEuRoparl-ST \cite{gaido-etal-2021-moby} & 37.7 & 30.1 & 26.0 \\
Triangle Multi \cite{gaido-etal-2022-talking} & 37.4 & 35.4 & 28.2 \\
\hline
Ours & \textbf{37.9} & \textbf{36.2} & \textbf{28.3} \\
\hline
\end{tabular}
\caption{BLEU ($\uparrow$) of our direct ST system in comparison with previous ST works on Europarl-ST.}
\label{tab:comparison_prev_work}
\end{table}

In Table \ref{tab:results}, instead, we compare
our cascade ST+NER and ASR+MT+NER baselines, the joint
ST\&NER inline and parallel methods, and the only previous work \cite{gaido-etal-2021-moby} that reports scores (NE accuracy) on
the NEuRoparl-ST benchmark
(using a direct ST system trained on a large amount of data).
%
We can notice that, even though trained on fewer data, 
the direct ST models of our cascade ST+NER baselines compare favourably with the previous work not only in terms of
translation quality (BLEU), 
but also in NE accuracy.
In particular, the NE accuracy of our cascade ST+NER baseline is superior on average on the three language pairs, as the gains in en-es (+0.5) and en-fr (+1.9) are only partially balanced by the small drop in en-it (-0.8). 
In addition, the full cascade model (ASR+MT+NER) -- despite leveraging NLLB, which is trained on a large amount of data, and the good ASR performance (12.5 WER, slightly better than the ASR trained on thousands of hours presented in \cite{gaido-etal-2021-moby}) -- is inferior to the ST+NER baseline on all metrics, with the only exception of the en-fr BLEU. 
This further demonstrates the strength of our cascade ST+NER baseline.

Focusing on
the comparison of the cascade and joint methods in performing the ST and NER tasks, we notice that 
the performance of both \textit{inline} and \textit{parallel} models are
close in terms of translation quality, both generic (BLEU) and specific to NEs (NE accuracy), compared to the ST+NER baseline.
The small differences among the scores of the various methods (up to 0.2 BLEU and up to 0.6 NE accuracy) are not consistent across language directions and are never statistically significant, 
thus being ascribable to fluctuations due to the inherent randomness of neural methods.
We can conclude that the additional NER task does not bring any improvement to ST in terms of NE translation
(in contrast with previous findings for MT \cite{xie-2022-e2e-ea-nmt})
but also does not degrade
translation quality, as it could have happened since part of the model 
is dedicated to the additional task.

\noindent\textbf{NE Recognition.} When we consider the F1 metric, instead, the results highlight the differences between the various approaches. Our joint NER and ST approaches beat the cascade by a statistically significant margin on all language pairs (0.4-1.0 F1). This is surprising if we consider that the training data of the joint methods was generated with the NER system of the cascade approach, and highlights the strength of direct multitask systems. Among the joint solutions, the \textit{inline} and \textit{parallel + NE emb.} significantly outperform the \textit{parallel} method, 
proving the importance of 
feeding the decoder with information about the NE category 
predicted for the previously generated tokens. 
The difference between \textit{inline} and \textit{parallel + NE emb.}, however, is
very 
small (0.1, if any)
and 
not statistically significant. These two methods can therefore be considered on par.

\begin{figure}[!t]
\includegraphics[width=0.46\textwidth]{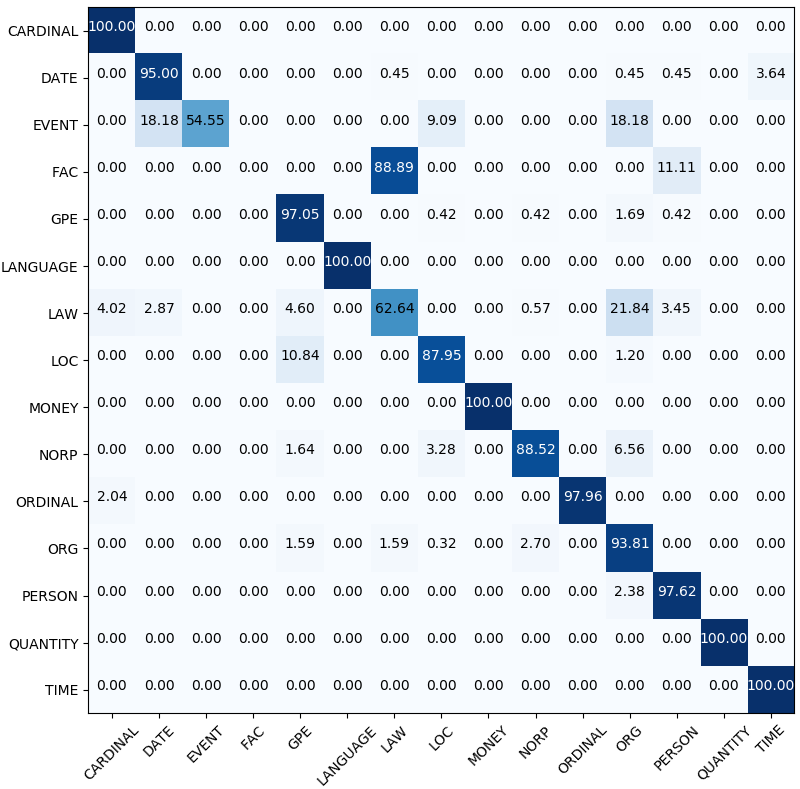}
\caption{\label{fig:confusion_matrix} Confusion matrix over the 15 NE categories with at least one NE correctly translated and recognized for the \textit{parallel + NE emb.} system on en-es. On the y-axis, there are the true labels, while on the x-axis the predicted labels. The numbers are percentages computed on the y-axis.} 
\end{figure}

\noindent\textbf{NE Classification.}
Lastly, all systems (joint and cascade) show a good ability in NE category classification.
The accuracy differences range between 0.6 and 0.3, are not coherent across language pairs and are never statistically significant. 
Not only their overall performance is on par, but also their confusion matrices over the NE categories are basically the same on all language pairs.
As an example, Fig. \ref{fig:confusion_matrix} 
reports
the confusion matrix of the \textit{parallel + NE emb.} model for en-es.
The classification accuracy is high (87.95-100\%) for all categories but three: facilities (\textit{FAC}), events (\textit{EVENT}), and names of laws (\textit{LAW}).
\textit{FAC} and \textit{EVENT} are very rare (19 and 9 occurrences in the test set), while \textit{LAW} is more frequent (141 occurrences), thus representing the main source of classification errors. The root of this difficulty may lay in the nature of law names, which have high variability, are long, and frequent only in specific domains.
At last, another common source of errors is
classifying \textit{GPE}
as location names, which is 
unsurprising as their categorization
highly depends on the context in which they occur (e.g. \textit{Europe} as a continent is a \textit{LOC}, but in politics it can 
be a
\textit{GPE}).

\section{Efficiency in Simultaneous ST}
\label{sec:efficiency}

One known advantage of direct systems over
cascade
ones is their lower overall computational cost since they need a forward pass on only one model instead of two.
For this reason, in applications where the computational cost is particularly critical, such as in simultaneous ST (SimulST), where 
it 
directly affects the output latency, direct ST systems obtain a significantly better latency-quality trade-off than ASR+MT solutions \cite{anastasopoulos-etal-2021-findings,anastasopoulos-etal-2022-findings}. 

However,
in our case of ST and NER, the computational cost is not only determined by the choice of a cascade 
or a full direct
system, but also by which of the two joint solutions is used. Indeed,
the number of decoding steps (i.e. forward passes on the autoregressive decoder)
required by the \textit{inline} and \textit{parallel} systems is different:
the former
method has to predict the start and end NE tags, requiring on average 7\% more decoding steps on the Europarl-ST test set
compared to a plain ST model and to the \textit{parallel} systems, which 
do
not introduce additional decoding steps.
%
%
%
For this reason, we conclude our work by comparing the two best models (\textit{inline} and \textit{parallel + NE emb}) in the simultaneous setting using the popular wait-k \cite{ma-etal-2020-simulmt} policy.

The wait-k policy consists in initially waiting for a predefined number of words~($k$) before starting to alternate between WRITE (emit a word) and READ (wait for more input audio) actions. 
Since the source is speech, the information about the number of words is not already present in the input, therefore a word detection strategy is applied to determine how many words have been pronounced at each time step. Here, we use
an
adaptive word detection strategy \cite{ren-etal-2020-simulspeech,zeng-etal-2021-realtrans}
that estimates the number of words in an audio segment by counting them in the
transcripts predicted by the CTC module trained on the encoded audio.
The choice of this method is motivated by its favorable
performance compared to
other word detection strategies \cite{zheng-etal-2020-simultaneous}. 
The wait-k policy is directly applied to offline-trained models without the need for any adaptation for the
task, as this approach has been demonstrated to be competitive with the one adopting models specifically trained to work in simultaneous \cite{papi-etal-2022-simultaneous}.

\pgfplotstableread[row sep=\\]{
k	BLEU	F1	LAAL	LAAL_CA \\
1	18.36	0.357	1374	2300 \\
2	25.42	0.414	1817	2905 \\
3	30.08	0.442	2193	3417 \\
}\inlinesim

\pgfplotstableread[row sep=\\]{
k	BLEU	F1	LAAL	LAAL_CA \\
1	19.07	0.365	1374	2249 \\
2	26.67	0.422	1817	2891 \\
3	31.36	0.456	2201	3374 \\
}\parallelsim

\pgfplotstableread[row sep=\\]{
k	BLEU	F1	LAAL	LAAL_CA \\
1	19.93	0	1366	2266 \\
2	26.30	0	1784	2841 \\
3	31.07	0	2163	3342 \\
}\onlySTsim

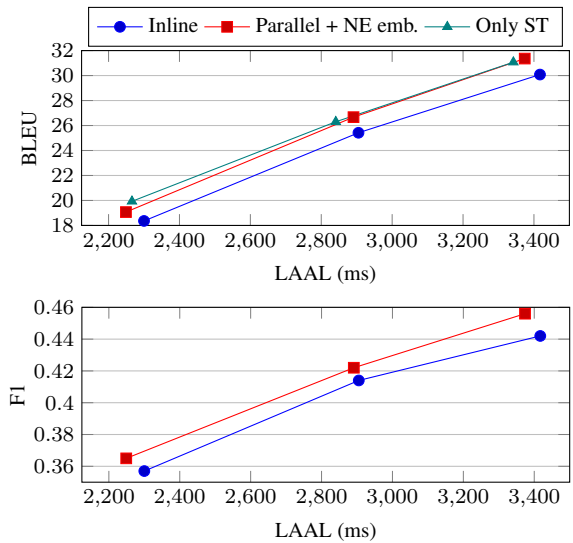
\begin{figure}[!t]
\centering
\quad
    \begin{subfigure}[t]{0.45\textwidth}
\begin{tikzpicture}
    \footnotesize
    \begin{axis}[
            ymajorgrids=true,
            ylabel=BLEU, xlabel=LAAL (ms),
            width=8cm,
            height=3.9cm,
            ymax=32, ymin=18,
            xmax=3500,
            ytick={18,20,22,24,26,28,30,32},
            legend style={at={(0.5,1.25)},
                anchor=north,legend columns=-1},
            xtick=data,
            xtick={},
        ]
        \addplot table[x=LAAL_CA,y=BLEU]{\inlinesim};
        \addplot table[x=LAAL_CA,y=BLEU]{\parallelsim};
        \addplot[color=teal, mark=triangle*] table[x=LAAL_CA,y=BLEU]{\onlySTsim};
        \legend{Inline, Parallel + NE emb., Only ST}
    \end{axis}
\end{tikzpicture}
    \end{subfigure}
\\
    \begin{subfigure}[t]{0.45\textwidth}
\begin{tikzpicture}
    \footnotesize
    \begin{axis}[
            ymajorgrids=true,
            ylabel=F1, xlabel=LAAL (ms),
            width=8cm,
            height=3.9cm,
            ymax=0.46,
            ymin=0.35,
            xmax=3500,
            ytick={0.36,0.38,0.40,0.42,0.44,0.46},
            legend style={at={(1.0,1.0)},
                anchor=south,legend columns=-1},
            xtick=data,
            xtick={},
            ylabel shift={-2pt},
        ]
        \addplot table[x=LAAL_CA,y=F1]{\inlinesim};
        \addplot table[x=LAAL_CA,y=F1]{\parallelsim};
    \end{axis}
\end{tikzpicture} 
    \end{subfigure}
\caption{\label{fig:sim} BLEU-
and F1-LAAL curves for en-es of the inline and parallel + NE emb solutions (we also include an ST-only system as reference).
Each point corresponds to a different value of $k=\{1,2,3\}$ and is
the average over three models.}
\end{figure}

Evaluating the performance in simultaneous
allows us to estimate the overhead introduced by the additional decoding steps of the \textit{inline} model compared to the \textit{parallel + NE emb.} one.
In Fig. \ref{fig:sim}, we report the BLEU-
and F1-latency curves computed on the outputs obtained by running the SimulEval tool \cite{ma-etal-2020-simuleval} on
the two joint NER\&ST models  
for en-es (for the sake of brevity, we do not report the curves for en-fr and en-it that show the same trends).
We also report the BLEU-latency curve of the direct ST-only model as a reference, while we do not show the cascade ST+NER as the computational cost (and hence, latency) is significantly higher.
Latency is measured through computational-aware length-adaptive average lagging (LAAL)~\cite{papi-etal-2022-generation}. The $k$ value of the wait-k policy is varied from 1 to 3 ($k=\{1,2,3\}$), in order to reach different latency regimes.

The curves show that the \textit{parallel + NE emb} model has the same latency and quality of an ST-only model, despite the additional NER task to perform. The \textit{inline} solution, instead, has similar quality but features a (slightly) increased latency, because of its higher computational cost.
However, since the computational cost only accounts for a fraction of the latency ($\sim$53\% of the computational-aware LAAL is due to the wait time of the wait-k policy), and the computational difference is not large ($\sim$5\%), the gap between the two models is limited.

All in all, we can conclude that the \textit{inline} model introduces a computational overhead that depends on the number of NEs detected in an utterance. 
On our test set, with 1,267 sentences, 30.6K words, and 1,638 NEs, we estimated as 5\% its computational overhead in time compared to a base direct ST model and to our \textit{parallel + NE emb.} solution.
In light of the similar quality of \textit{inline} and \textit{parallel + NE emb.} systems, this difference
-- 
which may be larger in domains where NEs are more frequent, as news or molecular biology \cite{nobata-etal-2000-comparison}
-- makes the \textit{parallel + NE emb.} method our best solution overall.

\section{Conclusions}
\label{sec:conclusions}

We presented the first multitask models jointly performing speech translation and named entity recognition. First, we showed the importance of properly feeding information about the previously predicted NE tags, as done in the \textit{inline} and \textit{parallel + NE emb.} models. Second, and most importantly, we showed
that our joint solutions consistently outperform a cascade system on the NER task (by 0.4-1.0 F1), while being on par in terms of translation quality.
Lastly, we evaluated the computational efficiency of our methods and demonstrated that the \textit{parallel + NE emb.} system, which does not introduce noticeable overhead with respect to a plain ST model, is more  efficient than the \textit{inline} method, besides being on par
in terms of translation and NER quality. 
As such, it represents the most attractive solution to
jointly perform ST and NER,
especially in the simultaneous scenario where its computational-aware latency is the same as a single model performing the ST task only.

\section{Acknowledgment}
This work is part of the project Smarter Interpreting (https://smarter-interpreting.eu/) financed by CDTI Neotec funds.
We acknowledge the support  of the PNRR project FAIR -  
Future AI Research (PE00000013),  under the NRRP MUR 
program funded by the NextGenerationEU.

\bibliographystyle{IEEEtran}
\bibliography{mybib}

\end{document}